\documentclass[11pt]{article}
\usepackage[a4paper,margin=25mm]{geometry}
\usepackage{amsmath,amssymb,bm}
\usepackage{booktabs,array}
\usepackage{placeins}
\usepackage{graphicx}
\usepackage[ruled,vlined,linesnumbered]{algorithm2e}
\usepackage{cite}
\usepackage[hidelinks]{hyperref}

\begin{document}

\title{Variational Mode Decomposition as Weighted Fuzzy $c$-Means}
\author{Yuya Ohmichi\\[2pt]
{\small Aviation Technology Directorate, Japan Aerospace Exploration Agency}\\
{\small Chofu, Tokyo 182-8522, Japan}}
\date{}
\maketitle

\begin{abstract}
Variational mode decomposition (VMD) decomposes a signal into modes of small bandwidth and is usually computed by the alternating direction method of multipliers (ADMM).
We show that its constrained variational problem is equivalent to weighted fuzzy $c$-means (FCM) with weighting exponent $m = 2$, with frequencies as data points and the power spectrum as their weights.
Its quadratic-penalty form corresponds to FCM with a noise cluster at distance $\delta = 1/\sqrt{2\alpha}$, where $\alpha$ is the penalty factor.
The ADMM fixed-point conditions coincide with the corresponding FCM membership and center conditions.
In the constrained form, these conditions are independent of $\alpha$ and the dual-ascent parameter.
The resulting FCM-based algorithm updates only the center frequencies during iteration, requires no dual-ascent parameter and requires $\alpha$ only in the quadratic-penalty form.
We extend the equivalence to multivariate VMD (MVMD) and two-dimensional VMD, with conjugate frequency pairs as data points in the latter.
For successive MVMD (SMVMD), we derive an FCM-based algorithm using FCM memberships to express the optimal mode for a fixed center at each extraction step.
For both MVMD and SMVMD, the computational cost per iteration of the FCM-based algorithm is independent of the number of channels.
In experiments with synthetic data, the FCM-based algorithms achieved results comparable to those of ADMM with fewer iterations and shorter computing times.
This equivalence provides a basis for applying the theory and methods developed for FCM to VMD.

\end{abstract}

\section{Introduction}
Variational mode decomposition (VMD) \cite{dragomiretskiy2014} decomposes a signal into modes of small bandwidth by solving a variational problem.
Each mode is modeled as a component concentrated around a center frequency.
Multivariate VMD (MVMD) \cite{rehman2019} extends VMD to multichannel signals with shared center frequencies, and two-dimensional VMD (2D-VMD) \cite{dragomiretskiy2015} extends VMD to images.
Successive MVMD (SMVMD) \cite{liu2022a} combines successive VMD \cite{nazari2020} with MVMD to extract modes successively without specifying their number in advance.

VMD is usually computed by the alternating direction method of multipliers (ADMM) \cite{boyd2011}, which updates the modes, center frequencies and a Lagrangian multiplier in turn.
Besides the number of modes $K$, the ADMM algorithm requires the penalty factor $\alpha$ and the dual-ascent parameter $\tau$.
The parameter $\alpha$ controls the bandwidth of the Wiener filter used in the mode update \cite{dragomiretskiy2014}, but its effect on the resulting decomposition is not fully understood.
Empirically, the decomposition has been reported to change little over a wide range of $\alpha$ in some cases \cite{eriksen2023,ohmichi2024}.

Fuzzy $c$-means (FCM) \cite{dunn1973,bezdek1981} is a clustering method that assigns each data point to the clusters with memberships between 0 and 1 that sum to 1 for each data point.
The FCM algorithm alternately updates the memberships and the cluster centers.
The noise cluster \cite{dave1991} accounts for data that do not belong to any of the regular clusters.
The FCM objective can be expressed in terms of the cluster centers alone \cite{hathaway1995}, and the convergence of the iteration has been analyzed \cite{bezdek1980,bezdek1987,hoppner2003,groll2005}.

Tang et al.~\cite{tang2025} incorporated FCM into MVMD by clustering the modes of each channel according to their contribution to the shared center frequencies, and solved the resulting variational model by ADMM.
Jia et al.~\cite{jia2026a} related VMD to weighted $k$-means by assuming a hard partition of the spectrum, in which each frequency is assigned to a single mode.
In the variational problem of VMD, however, the spectrum at each frequency can be shared by several modes.
For MVMD, Ohmichi~\cite{ohmichi2026} showed that, at an ADMM fixed point with $\tau > 0$, the spectrum is distributed among the modes in inverse proportion to the squared distances from the center frequencies, independently of $\alpha$.
Jia et al.~\cite{jia2026b} also derived the spectral distribution at ADMM fixed points and identified the inverse-square distribution as the minimizer of the bandwidth term for fixed centers.
They further introduced a regularized variational problem that yields a soft spectral partition with Lorentzian weights.

This paper shows that the constrained variational problem of VMD is equivalent to weighted FCM with weighting exponent $m = 2$, where the frequencies are the data points and the power spectrum gives their weights.
Its quadratic-penalty form corresponds to FCM with a noise cluster at distance $\delta = 1/\sqrt{2\alpha}$.
The ADMM fixed-point conditions coincide with the membership and center conditions of the corresponding FCM problems.
In the constrained form, the ADMM fixed-point conditions for the modes and centers are independent of $\alpha$ and $\tau$.
Based on this equivalence, we derive an FCM-based algorithm for VMD that updates only the center frequencies during the iteration.
The algorithm requires no dual-ascent parameter $\tau$ and requires $\alpha$ only in the quadratic-penalty form.

We extend the equivalence to MVMD and 2D-VMD and derive their FCM-based algorithms.
We also derive an FCM-based algorithm for SMVMD by expressing the optimal mode for a fixed center at each extraction step using FCM memberships.
For both MVMD and SMVMD, the computational cost per iteration of the FCM-based algorithm is independent of the number of channels, whereas that of ADMM grows with the number of channels.
Numerical experiments compare these algorithms with the corresponding ADMM algorithms in terms of decomposition results and computational cost.

\section{VMD and FCM}\label{sec:vmd}
\subsection{VMD and Its Frequency-Domain Form}\label{sec:vmdform}
VMD \cite{dragomiretskiy2014} decomposes a real-valued input signal $f(t)$ into $K$ real-valued modes $u_k(t)$ with center angular frequencies $\omega_k$ by solving
\begin{equation}
\min_{\{u_k\},\{\omega_k\}} \sum_{k} \bigl\|\partial_t \bigl[u_k^{+}(t)\,e^{-j\omega_k t}\bigr]\bigr\|_2^2
\ \ \text{s.t.}\ \ \sum_{k} u_k = f,\label{eq:time}
\end{equation}
where $u_k^{+} = u_k + j\mathcal{H} u_k$ is the analytic signal of $u_k$ and $\mathcal{H}$ is the Hilbert transform.
Each term penalizes the spectral spread of a mode around its center frequency, and the constraint requires the modes to reconstruct the signal.

To solve \eqref{eq:time} by ADMM, \cite{dragomiretskiy2014} forms the augmented Lagrangian
\begin{equation}
\alpha\sum_k \bigl\|\partial_t \bigl[u_k^{+} e^{-j\omega_k t}\bigr]\bigr\|_2^2 + \Bigl\|f - \sum_k u_k\Bigr\|_2^2 + \Bigl\langle\lambda,f - \sum_k u_k\Bigr\rangle\label{eq:al}
\end{equation}
with the penalty factor $\alpha > 0$ and the Lagrangian multiplier $\lambda(t)$.
The penalty factor $\alpha$ weights the bandwidth term against the squared reconstruction error and is absent from the constrained problem \eqref{eq:time} itself.

Let $\hat u_k(\omega)$ be the Fourier transform of $u_k$.
The spectrum of the analytic signal is $\hat u_k^{+}(\omega) = 2\hat u_k(\omega)$ for $\omega > 0$ and $\hat u_k^{+}(\omega) = 0$ for $\omega < 0$.
Multiplication by $e^{-j\omega_k t}$ shifts this spectrum to $\hat u_k^{+}(\omega + \omega_k)$, and $\partial_t$ multiplies it by $j\omega$.
By Parseval's identity, each term of \eqref{eq:time} is therefore
\begin{equation}
\begin{aligned}
\bigl\|\partial_t \bigl[u_k^{+} e^{-j\omega_k t}\bigr]\bigr\|_2^2
&= \frac{1}{2\pi} \int_{-\infty}^{\infty} \omega^2|\hat u_k^{+}(\omega + \omega_k)|^2\,d\omega\\
&= \frac{2}{\pi} \int_0^\infty(\omega - \omega_k)^2|\hat u_k(\omega)|^2\,d\omega.
\end{aligned}\label{eq:parseval}
\end{equation}

Because $u_k$ is real, $\hat u_k(-\omega)$ is the complex conjugate of $\hat u_k(\omega)$, so the spectrum on $\omega \ge 0$ determines the mode.
For a sampled real signal $f$ of length $T$, let $\Omega$ be the set of discrete angular frequencies $\omega_n = 2\pi n/T$ (rad/sample) with $n = 0,\ldots,\lceil T/2\rceil - 1$.
Let $M = |\Omega|$ and $\Delta\omega = 2\pi/T$ be the number of frequencies and the frequency resolution, respectively.
We denote the unilateral spectra of the input and modes by $\hat f(\omega)$ and $\hat u_k(\omega)$, and the input power spectrum by $P(\omega) = |\hat f(\omega)|^2$.
All sums over $\omega$ are taken over $\Omega$.
The discrete counterpart of \eqref{eq:time} can then be written, up to a constant factor, as
\begin{equation}
\begin{aligned}
\min_{\{\hat u_k\},\{\omega_k\}}\ & \sum_{\omega\in\Omega} \sum_{k=1}^{K}(\omega - \omega_k)^2|\hat u_k(\omega)|^2\\
\text{s.t.}\ \ & \sum_{k=1}^{K} \hat u_k(\omega) = \hat f(\omega) \quad(\forall\omega \in \Omega).
\end{aligned}\label{eq:vmd}
\end{equation}

\subsection{Equivalence to Weighted FCM}\label{sec:equiv}
We show that \eqref{eq:vmd} is equivalent to weighted FCM \cite{dunn1973,bezdek1981,havens2012} with weighting exponent $m = 2$.
For scalar data points $x_n$ with weights $w_n \ge 0$, this problem seeks memberships $\mu_{kn}$ and cluster centers $v_k$ that solve
\begin{equation}
\begin{aligned}
\min_{\{\mu_{kn}\},\{v_k\}}\ & \sum_{n} w_n \sum_{k=1}^{K} \mu_{kn}^m(x_n - v_k)^2\\
\text{s.t.}\ \ & \mu_{kn} \ge 0,\quad\sum_{k=1}^{K} \mu_{kn} = 1\quad(\forall n).
\end{aligned}\label{eq:fcm}
\end{equation}
Here, $m > 1$ is the weighting exponent.
For $m = 2$, the optimal memberships for fixed cluster centers and the optimal cluster centers for fixed memberships are given by
\begin{equation}
\mu_{kn} = \frac{(x_n - v_k)^{-2}}{\sum_{\ell=1}^{K}(x_n - v_\ell)^{-2}},\qquad
v_k = \frac{\sum_n w_n \mu_{kn}^2 x_n}{\sum_n w_n \mu_{kn}^2},\label{eq:fcm_update}
\end{equation}
and the FCM iteration alternates these two updates \cite{bezdek1981}.

For fixed center frequencies, \eqref{eq:vmd} separates over the frequencies.
At one frequency $\omega$, with $d_k = |\omega - \omega_k|$, we minimize $\sum_k d_k^2|\hat u_k|^2$ over the complex numbers $\hat u_k$ subject to $\sum_k \hat u_k = \hat f$.
If $d_k > 0$ for all $k$, this is a strictly convex quadratic problem with the Lagrangian
\begin{equation}
\sum_{k} d_k^2|\hat u_k|^2 - 2\,\mathrm{Re}\Bigl[\nu^{*} \Bigl(\sum_{k} \hat u_k - \hat f\Bigr)\Bigr],
\end{equation}
where $\nu^{*}$ is the complex conjugate of the multiplier $\nu(\omega)$.
Stationarity with respect to $\hat u_k$ gives $d_k^2 \hat u_k = \nu$.
Hence $\hat u_k = \nu d_k^{-2}$, and the constraint gives $\nu = \hat f/\sum_\ell d_\ell^{-2}$.
The unique minimizer is
\begin{equation}
\hat u_k = s_k \hat f,\qquad s_k = \frac{(\omega - \omega_k)^{-2}}{\displaystyle\sum_{\ell=1}^{K}(\omega - \omega_\ell)^{-2}}.\label{eq:member}
\end{equation}
Here $s_k$ is real, $s_k \ge 0$ and $\sum_k s_k = 1$, so every optimal mode is in phase with $\hat f$.
The coefficients $s_k$ coincide with the optimal FCM memberships in \eqref{eq:fcm_update}.
The same spectral distribution was also obtained in \cite{ohmichi2026,jia2026b}.

For fixed modes, the objective function of \eqref{eq:vmd} is quadratic in each $\omega_k$, and its minimizer is $\omega_k = \sum_\omega \omega|\hat u_k(\omega)|^2/\allowbreak\sum_\omega|\hat u_k(\omega)|^2$.
With the optimal modes \eqref{eq:member}, this becomes
\begin{equation}
\omega_k = \frac{\sum_{\omega} \omega P(\omega) s_k(\omega)^2}{\sum_{\omega} P(\omega) s_k(\omega)^2},\label{eq:center}
\end{equation}
which is the center update of FCM in \eqref{eq:fcm_update}.

With the optimal modes $\hat u_k = s_k \hat f$, the objective function of \eqref{eq:vmd} becomes $\sum_\omega P(\omega) \allowbreak\sum_k s_k(\omega)^2(\omega - \omega_k)^2$.
This is the FCM objective function in \eqref{eq:fcm} for $m = 2$, with the frequencies $\omega$ as the data points $x_n$, the power $P(\omega)$ as the weights $w_n$, the center frequencies $\omega_k$ as the centers $v_k$, and $s_k(\omega)$ as the memberships $\mu_{kn}$.
The form $\hat u_k = s_k \hat f$ is not imposed in the VMD problem but follows from optimality, as shown in \eqref{eq:member} when all distances are positive.
Since this correspondence holds for any fixed set of center frequencies, \eqref{eq:vmd} and \eqref{eq:fcm} with $m = 2$ have the same minimum objective value when the centers are optimized.

As in FCM, the VMD objective can be expressed in terms of the center frequencies alone.
Substituting the optimal modes \eqref{eq:member} into \eqref{eq:vmd} yields the reduced objective function
\begin{equation}
\begin{aligned}
J_0(\{\omega_k\}) &= \sum_{\omega} P(\omega) \sum_{k} s_k(\omega)^2 d_k(\omega)^2\\
&= \sum_{\omega} \frac{P(\omega)}{\sum_k d_k(\omega)^{-2}},
\end{aligned}\label{eq:J0}
\end{equation}
where the second equality follows from $\sum_k s_k^2 d_k^2 = \allowbreak 1/\allowbreak\sum_k d_k^{-2}$.
This is the reformulated FCM objective function of \cite{hathaway1995}; when $P(\omega) d_k(\omega)^2 > 0$ for all $\omega$ and $k$, its local minimizers correspond to those of the original FCM \cite{hathaway1995}.

\subsection{Quadratic Penalty and the Noise Cluster}\label{sec:pen}
Removing the Lagrangian multiplier term from \eqref{eq:al} gives the quadratic-penalty form of VMD \cite{dragomiretskiy2014}, in which the reconstruction constraint is replaced by a quadratic penalty.
By Parseval's identity, as in \eqref{eq:parseval}, the squared reconstruction error is $\|f - \sum_k u_k\|_2^2 = (1/\pi) \int_0^\infty|\hat f - \sum_k \hat u_k|^2\,d\omega$.
Dividing the objective function by $2\alpha/\pi$ yields, on the discrete frequencies,
\begin{equation}
\min_{\{\hat u_k\},\{\omega_k\}} \sum_{\omega\in\Omega} \Bigl[\sum_k d_k(\omega)^2|\hat u_k(\omega)|^2 + \frac{1}{\gamma} \Bigl|\hat f(\omega) - \sum_k \hat u_k(\omega)\Bigr|^2\Bigr],\label{eq:pen}
\end{equation}
with $\gamma = 2\alpha > 0$.

The quadratic-penalty form \eqref{eq:pen} is equivalent to FCM with a noise cluster \cite{dave1991}.
The noise cluster adds a cluster $k = 0$ to \eqref{eq:fcm} with $m = 2$ at a constant distance $\delta$, the noise distance from every data point:
\begin{equation}
\begin{aligned}
\min_{\{\mu_{kn}\},\{v_k\}}\ & \sum_{n} w_n \Bigl[\sum_{k=1}^{K} \mu_{kn}^2(x_n - v_k)^2 + \delta^2 \mu_{0n}^2\Bigr]\\
\text{s.t.}\ \ & \mu_{kn} \ge 0,\quad\sum_{k=0}^{K} \mu_{kn} = 1\quad(\forall n).
\end{aligned}\label{eq:noise_fcm}
\end{equation}

Defining the residual as $\hat u_0 = \hat f - \sum_k \hat u_k$, we can rewrite \eqref{eq:pen} in the form of \eqref{eq:vmd} with an additional component $\hat u_0$ weighted by $1/\gamma$ and the constraint $\sum_{k=0}^{K} \hat u_k = \hat f$.
For fixed centers, the optimality condition is $d_k^2 \hat u_k = \hat u_0/\gamma = \nu$.
Hence $\hat u_k = \nu d_k^{-2}$ and $\hat u_0 = \gamma\nu$, and the constraint gives $\nu = \hat f/(\gamma + \sum_\ell d_\ell^{-2})$.
As in \eqref{eq:member}, we obtain
\begin{equation}
\begin{aligned}
s_k &= \frac{d_k^{-2}}{\gamma + \sum_\ell d_\ell^{-2}}\ (k \ge 1),\quad s_0 = \frac{\gamma}{\gamma + \sum_\ell d_\ell^{-2}},\\
\hat u_k &= s_k \hat f,\qquad \hat f - \textstyle\sum_k \hat u_k = s_0 \hat f.
\end{aligned}\label{eq:noise}
\end{equation}

With $\hat u_k = s_k \hat f$ and $\hat u_0 = s_0 \hat f$, the objective function of \eqref{eq:pen} becomes $\sum_\omega P(\omega) \allowbreak[\sum_k s_k(\omega)^2 d_k(\omega)^2 + s_0(\omega)^2/\gamma]$.
This coincides with the FCM objective in \eqref{eq:noise_fcm} when the noise distance is
\begin{equation}
\delta = \frac{1}{\sqrt\gamma} = \frac{1}{\sqrt{2\alpha}},\label{eq:delta}
\end{equation}
and the coefficients $s_k$ and $s_0$ in \eqref{eq:noise} coincide with the optimal FCM memberships under the correspondence established in Section~\ref{sec:equiv}.
Thus, the quadratic-penalty form of VMD is equivalent to FCM with a noise cluster whose distance is determined by $\alpha$.
By \eqref{eq:noise}, $s_0/s_k = \gamma d_k^2$, so the membership in the noise cluster exceeds that in mode $k$ exactly when the distance between $\omega$ and $\omega_k$ exceeds $\delta$.

\subsection{Unified Reduced Objective and Stationarity Conditions}
We now extend the reduced objective function $J_0$ in \eqref{eq:J0} to the quadratic-penalty form and derive the stationarity conditions common to both forms.
Substituting \eqref{eq:noise} into \eqref{eq:pen} and using $\sum_k s_k^2 d_k^2 + s_0^2/\gamma = 1/(\gamma + \sum_\ell d_\ell^{-2})$ yields the unified reduced objective function
\begin{equation}
J_\gamma(\{\omega_k\}) = \sum_{\omega\in\Omega} \frac{P(\omega)}{\gamma + \sum_{k=1}^{K}(\omega - \omega_k)^{-2}} \qquad(\gamma \ge 0),\label{eq:Jgamma}
\end{equation}
where $\gamma = 0$ recovers the constrained form \eqref{eq:J0} and $\gamma > 0$ corresponds to the quadratic-penalty form.
Differentiating \eqref{eq:Jgamma} gives
\begin{equation}
\frac{\partial J_\gamma}{\partial\omega_k} = -2\sum_{\omega} P(\omega) s_k(\omega)^2(\omega - \omega_k).\label{eq:grad}
\end{equation}
Setting $\partial J_\gamma/\partial\omega_k = 0$ in \eqref{eq:grad} and rearranging yields the center condition \eqref{eq:center}, with the memberships given by \eqref{eq:noise} for $\gamma > 0$ and \eqref{eq:member} for $\gamma = 0$.

\subsection{ADMM Fixed Points and FCM Conditions}\label{sec:admm}
We now show that the fixed-point conditions of ADMM reduce to the membership and center conditions of FCM.
The ADMM updates of VMD \cite{dragomiretskiy2014}, based on the augmented Lagrangian \eqref{eq:al}, are written on the unilateral spectrum as
\begin{equation}
\hat u_k \leftarrow \frac{\hat f - \sum_{\ell\ne k} \hat u_\ell + \hat\lambda/2}{1 + 2\alpha(\omega - \omega_k)^2},\qquad
\omega_k \leftarrow \frac{\sum_\omega \omega|\hat u_k(\omega)|^2}{\sum_\omega|\hat u_k(\omega)|^2},\label{eq:admm}
\end{equation}
\begin{equation}
\hat\lambda \leftarrow \hat\lambda + \tau\Bigl(\hat f - \sum_k \hat u_k\Bigr),\label{eq:dual}
\end{equation}
where $\hat\lambda$ is the Fourier transform of $\lambda$, $\tau \ge 0$ is the dual-ascent parameter, and $\hat u_k$ is updated in the order of $k$.

At a fixed point, the mode update in \eqref{eq:admm} leaves every $\hat u_k$ unchanged.
Multiplying it by the denominator and subtracting $\hat u_k$ from both sides gives
\begin{equation}
2\alpha d_k^2 \hat u_k = \hat f - \sum_{\ell=1}^{K} \hat u_\ell + \frac{\hat\lambda}{2} \qquad(k = 1,\ldots,K).\label{eq:fixed_point}
\end{equation}

\emph{Case $\tau > 0$:} At a fixed point, \eqref{eq:dual} gives $\sum_k \hat u_k = \hat f$, and \eqref{eq:fixed_point} reduces to $d_k^2 \hat u_k = \hat\lambda/(4\alpha)$.
This is the optimality condition of Section~\ref{sec:equiv} with $\nu = \hat\lambda/(4\alpha)$.
Together with the reconstruction constraint, it yields the membership condition \eqref{eq:member}.
The center update in \eqref{eq:admm} reduces to the center condition \eqref{eq:center}, equivalent to $\nabla J_0 = 0$ by \eqref{eq:grad}.
Conversely, for any stationary point of $J_0$, choosing the modes as in \eqref{eq:member} and $\hat\lambda = 4\alpha\hat f/\sum_\ell d_\ell^{-2}$ leaves all variables unchanged under the ADMM updates \eqref{eq:admm} and \eqref{eq:dual}.
Thus, the ADMM fixed-point conditions for the centers and modes are equivalent to those of FCM and involve neither $\alpha$ nor $\tau$.

\emph{Case $\tau = 0$:} The multiplier $\hat\lambda$ remains unchanged by \eqref{eq:dual} and is therefore zero if initialized to zero.
Equation \eqref{eq:fixed_point} then reduces to $d_k^2 \hat u_k = \hat u_0/(2\alpha)$, which is the optimality condition of the quadratic-penalty form \eqref{eq:pen} with $\gamma = 2\alpha$.
The modes are therefore given by \eqref{eq:noise}, and the center update reduces to \eqref{eq:center}, equivalent to $\nabla J_{2\alpha} = 0$ by \eqref{eq:grad}.
Conversely, for any stationary point of $J_{2\alpha}$, choosing the modes as in \eqref{eq:noise} and setting $\hat\lambda = 0$ leaves all variables unchanged under the ADMM updates \eqref{eq:admm} and \eqref{eq:dual}.
Thus, the fixed-point conditions are equivalent to those of FCM with a noise cluster and depend on $\alpha$ through the noise distance \eqref{eq:delta}.

The conditions above concern only the fixed points of ADMM, not its convergence.
Because the problem is nonconvex, ADMM is not guaranteed to converge \cite{boyd2011}.

\subsection{FCM-Based Algorithm for VMD}
Based on the equivalence, VMD can be computed by an FCM-based algorithm, given as Algorithm~\ref{alg:vmd}, which applies the FCM iteration \cite{bezdek1981}.
The algorithm is written for $Q$ channels so that it also computes MVMD (Section~\ref{sec:mvmd}); for VMD, $Q = 1$ and $P_{\mathrm{tot}} = P$.
Each iteration computes the memberships from the current center frequencies and then updates the centers.
The modes are constructed only once, after the iteration terminates.
In contrast, ADMM updates the modes, center frequencies and Lagrangian multiplier in every iteration.

Algorithm~\ref{alg:vmd} solves the constrained form \eqref{eq:vmd} for $\gamma = 0$ and the quadratic-penalty form \eqref{eq:pen} for $\gamma > 0$.
In the latter case, the only model parameter besides $K$ is the noise distance $\delta = 1/\sqrt\gamma$.
The membership update step evaluates \eqref{eq:noise}, which reduces to \eqref{eq:member} for $\gamma = 0$.
If one or more centers coincide with a discrete frequency, the membership at that frequency is divided equally among them and all other memberships are set to zero, following the FCM rule for zero distances \cite{bezdek1981}.
Here, $[a_k = 0]$ in Algorithm~\ref{alg:vmd} equals 1 if $a_k = 0$ and 0 otherwise.

\begin{algorithm}[t]
  \DontPrintSemicolon
  \SetKwComment{Comment}{$\triangleright$\ }{}
  \SetCommentSty{textnormal}
  \caption{FCM-based algorithm for VMD and MVMD}\label{alg:vmd}
  \KwIn{unilateral spectra $\hat f_q(\omega)$, $q = 1,\ldots,Q$, $\Omega$, $K$, $\gamma \ge 0$, initial centers $\omega_k \in [0,\pi]$, tolerance $\varepsilon$, maximum iterations $N$}
  \KwOut{centers $\omega_k$, modes $\hat u_{k,q}(\omega)$, residuals $\hat u_{0,q}(\omega)$}
  \For{each $\omega \in \Omega$}{
    $P_{\mathrm{tot}}(\omega) \gets \sum_{q=1}^{Q}|\hat f_q(\omega)|^2$\;
  }
  \For{$i = 1,\ldots,N$}{
    \For(\Comment*[f]{memberships}){each $\omega \in \Omega$}{
      \For{$k = 1,\ldots,K$}{
        $a_k \gets (\omega - \omega_k)^2$\;
      }
      \eIf{$\min_k a_k > 0$}{
        \For{$k = 1,\ldots,K$}{
          $s_k(\omega) \gets a_k^{-1} \big/\bigl(\gamma + \sum_\ell a_\ell^{-1}\bigr)$\;
        }
        $s_0(\omega) \gets \gamma\big/\bigl(\gamma + \sum_\ell a_\ell^{-1}\bigr)$\;
      }{
        $n_0 \gets$ number of $k$ with $a_k = 0$\;
        \For{$k = 1,\ldots,K$}{
          $s_k(\omega) \gets [a_k = 0]/n_0$\;
        }
        $s_0(\omega) \gets 0$\;
      }
    }
    \For(\Comment*[f]{centers}){$k = 1,\ldots,K$}{
      $\bar\omega_k \gets \dfrac{\sum_\omega \omega P_{\mathrm{tot}}(\omega) s_k(\omega)^2}{\sum_\omega P_{\mathrm{tot}}(\omega) s_k(\omega)^2}$\;
    }
    $r \gets \max_k|\bar\omega_k - \omega_k|/\Delta\omega$\;
    \lIf{$r < \varepsilon$ or $i = N$}{\textbf{break}}
    \For{$k = 1,\ldots,K$}{
      $\omega_k \gets \bar\omega_k$\;
    }
  }
  \For(\Comment*[f]{modes and residuals}){$q = 1,\ldots,Q$}{
    \For{$k = 0,\ldots,K$}{
      \For{each $\omega \in \Omega$}{
        $\hat u_{k,q}(\omega) \gets s_k(\omega) \hat f_q(\omega)$\;
      }
    }
  }
\end{algorithm}

The center update step evaluates the right-hand side of \eqref{eq:center}, with $P$ replaced by $P_{\mathrm{tot}}$, to obtain $\bar\omega_k$.
The iteration stops when the stationarity residual
\begin{equation}
r = \max_k \frac{|\omega_k - \bar\omega_k|}{\Delta\omega} \label{eq:residual}
\end{equation}
falls below the tolerance $\varepsilon$ or the maximum number of iterations is reached.
By \eqref{eq:grad}, $r = 0$ is equivalent to $\nabla J_\gamma = 0$.
The residual can be evaluated from the centers alone and thus provides a common stopping criterion for the FCM-based algorithm and ADMM.
Its evaluation costs $O(KM)$.

The membership and center update steps alternately minimize the FCM objective over their respective variables, so $J_\gamma$ does not increase along the iteration \cite{bezdek1980}.
The problem is nonconvex, and the point reached may depend on the initial centers.
Each iteration of Algorithm~\ref{alg:vmd} costs $O(KM)$, as does each ADMM iteration.

\subsection{Numerical Experiments}\label{sec:vmdexp}
We check that the points reached by ADMM satisfy the membership and center conditions of FCM and compare the iteration counts, computing times and objective values of ADMM and the FCM-based algorithm.

\emph{Test signals.} We use two single-channel test signals of 1000 samples, each consisting of three components, at a sampling frequency $f_{\mathrm{s}} = 1000$~Hz.
Both signals are decomposed with $K = 3$ after mirror extension by 500 samples at each end ($T = 2000$; $\Delta\omega$ corresponds to 0.5~Hz).
The first signal is a sum of sinusoids $\sum_k \cos(2\pi\xi_k t + \phi_k)$ with $\xi_k = 60.3$, 180.7 and 330.2~Hz.
The second, the AM--FM signal, is a sum of amplitude- and frequency-modulated signals $\sum_k(1 + 0.5\cos(2\pi 4t + \theta_k)) \cos(2\pi\xi_k t + 2\sin(2\pi 5t + \psi_k) + \phi_k)$.
The bandwidth of each component of the AM--FM signal is about 30~Hz.
The phases $\phi_k$, $\theta_k$ and $\psi_k$ are independent and uniform on $[0,2\pi)$.
White Gaussian noise is added at a signal-to-noise ratio (SNR) of 20 or 5~dB, where the SNR is the ratio of the sample variance of the noise-free signal to the noise variance.

\emph{Solvers.} ADMM and the FCM-based algorithm (Algorithm~\ref{alg:vmd}) are applied to both forms of VMD.
In the quadratic-penalty form, the noise distance is set to $\delta = 40$~Hz ($0.08\pi$~rad/sample), which is larger than the bandwidth of each component of the AM--FM signal.
The FCM-based algorithm uses $\gamma = 1/\delta^2$, and ADMM uses $\alpha = 1/(2\delta^2)$ and $\tau = 0$.
Both parameters of ADMM are thus fixed.

In the constrained form, the FCM-based algorithm uses $\gamma = 0$, whereas ADMM has two parameters, $\alpha$ and $\tau$, to be tuned.
The candidate values of $\alpha$ are specified through \eqref{eq:delta} by $\delta$ ranging from 10 to 640~Hz, and those of $\tau$ range from 0.1 to 5.
Each setting is evaluated with ten realizations of each case.
Among the settings for which all runs reached a stationary point within 1000 iterations, $\delta = 80$~Hz and $\tau = 3$ gave the smallest median number of iterations and are used in the following experiments.

Both solvers start from the same initial centers, the midpoints of the $K$ equal subintervals of $[0,\pi]$.
ADMM starts from $\hat u_k = 0$ and $\hat\lambda = 0$.

\emph{Measurement.} Each case is evaluated with 100 realizations, with the phases and noise regenerated for each realization.
For both solvers, the stationarity residual $r$ in \eqref{eq:residual} is evaluated after each iteration.
A stationary point is deemed to be reached when $r$ first falls below $10^{-6}$, at which point the iteration count and computing time are recorded.
Throughout this paper, all solvers run in a single thread on an AMD EPYC 9474F CPU, and the computing time includes the entire decomposition process, excluding the forward and inverse Fourier transforms, which are common to the compared solvers.
Objective values are normalized by the total power and reported as $J_\gamma/\sum_\omega P(\omega)$.

\begin{figure}[t]
\centering
\includegraphics{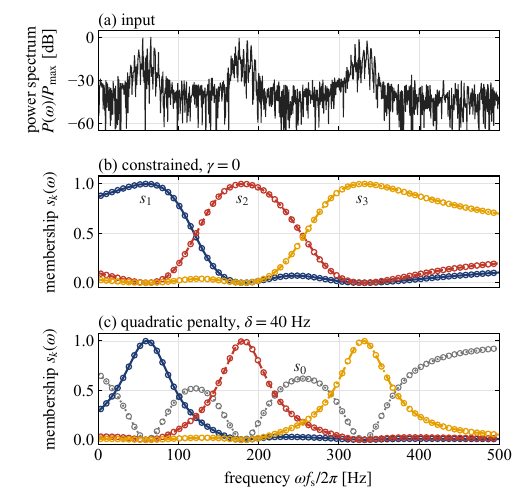}
\caption{Comparison of ADMM mode spectra and FCM memberships (AM--FM signal, SNR 20~dB).
(a) Input power spectrum $P(\omega)$ relative to its maximum $P_{\max}$.
(b) Constrained form.
(c) Quadratic-penalty form with $\delta = 40$~Hz.
Lines: FCM memberships $s_k(\omega)$ at the center frequencies reached by ADMM.
Circles: spectral amplitude ratios $|\hat u_k(\omega)|/|\hat f(\omega)|$ obtained by ADMM (every 14th discrete frequency).
Gray denotes the noise-cluster membership $s_0$ and the corresponding residual ratio.}\label{fig:membership}
\end{figure}

\emph{Verification of the fixed-point conditions.} Fig.~\ref{fig:membership} illustrates the agreement between the modes computed by ADMM and the FCM memberships for one realization of the AM--FM signal.
The ratios $|\hat u_k(\omega)|/|\hat f(\omega)|$ obtained by ADMM (circles) agree with the FCM memberships (lines).
In the constrained form, the whole spectrum is divided among the three modes.
In the quadratic-penalty form, the spectrum between and outside the three components is largely assigned to the noise cluster and forms the residual.

This agreement was confirmed in all 100 realizations of each case.
ADMM satisfied the center condition \eqref{eq:center} with $r < 10^{-6}$, and its modes agreed with those constructed from its center frequencies using \eqref{eq:member} and \eqref{eq:noise}, with a relative difference $\|\hat u - s\hat f\|/\|\hat f\| < 1.8 \times 10^{-7}$ over all modes and frequencies.

\begin{table}[t]
\caption{Median Iteration Counts and Computing Times to Reach a Stationary Point in VMD}\label{tab:vmd}
\centering
\begin{tabular}{@{}lllrrrr@{}}
\toprule
& & & \multicolumn{2}{c}{Iterations} & \multicolumn{2}{c}{Time [ms]}\\
\cmidrule(lr){4-5}\cmidrule(l){6-7}
Form & Signal & SNR & FCM & ADMM & FCM & ADMM\\
\midrule
Quadratic & Sinusoids & 20 dB & 6 & 10 & 0.15 & 0.60\\
penalty & Sinusoids & 5 dB & 12 & 15 & 0.29 & 0.90\\
& AM--FM & 20 dB & 11 & 15 & 0.27 & 0.90\\
& AM--FM & 5 dB & 16 & 20 & 0.38 & 1.18\\
\addlinespace
Constrained & Sinusoids & 20 dB & 6 & 29 & 0.14 & 1.71\\
& Sinusoids & 5 dB & 12 & 34 & 0.28 & 2.00\\
& AM--FM & 20 dB & 7 & 29 & 0.16 & 1.72\\
& AM--FM & 5 dB & 13 & 34 & 0.30 & 2.00\\
\bottomrule
\end{tabular}
\end{table}

\emph{Computational cost.} Table~\ref{tab:vmd} compares the median iteration counts and computing times required to reach a stationary point.
Both solvers reached a stationary point in all 100 realizations of each case.
In the quadratic-penalty form, ADMM needed 1.3--1.7 times as many iterations as the FCM-based algorithm and took 3.1--4.0 times as long.
In the constrained form, even with tuned parameters, ADMM needed 2.6--4.8 times as many iterations and took 6.7--12 times as long.
In each FCM iteration, the modes are optimal for the current centers, whereas ADMM updates each mode from the current values of the other modes and the Lagrangian multiplier.

\emph{Objective values.} In 798 of the 800 pairs of runs, the normalized objective values reached by the two solvers differed by less than $10^{-6}$ in relative terms.
The remaining two cases occurred in the constrained form at an SNR of 5~dB, one for the sinusoidal signal and one for the AM--FM signal.
In these cases, ADMM reached a different stationary point with an objective value 2.2--2.3 times that reached by the FCM-based algorithm.

\section{MVMD}\label{sec:mvmd}
MVMD \cite{rehman2019} extends VMD to multichannel signals.
We show that MVMD is also equivalent to weighted FCM and derive the corresponding FCM-based algorithm.

\subsection{Variational Problem and Equivalence to Weighted FCM}
MVMD \cite{rehman2019} decomposes $Q$ real-valued signals $f_q(t)$, $q = 1,\ldots,Q$, into $K$ modes per channel, $u_{k,q}(t)$, with center frequencies $\omega_k$ shared among the channels, by solving
\begin{equation}
\begin{aligned}
\min_{\{u_{k,q}\},\{\omega_k\}}\ & \sum_{k} \sum_{q} \bigl\|\partial_t \bigl[u_{k,q}^{+}(t)\,e^{-j\omega_k t}\bigr]\bigr\|_2^2\\
\text{s.t.}\ \ & \sum_{k} u_{k,q} = f_q \quad(\forall q),
\end{aligned}
\end{equation}
that is, \eqref{eq:time} summed over the channels with the reconstruction constraint imposed separately for each channel.
As in Section~\ref{sec:vmdform}, the discrete counterpart is
\begin{equation}
\begin{aligned}
\min_{\{\hat u_{k,q}\},\,\{\omega_k\}}\ & \sum_{q=1}^{Q} \sum_{\omega\in\Omega} \sum_{k=1}^{K}(\omega - \omega_k)^2|\hat u_{k,q}(\omega)|^2\\
\text{s.t.}\ \ & \sum_{k=1}^{K} \hat u_{k,q}(\omega) = \hat f_q(\omega) \quad(\forall\omega \in \Omega,\ \forall q).
\end{aligned}\label{eq:mvmd}
\end{equation}

For fixed centers, \eqref{eq:mvmd} and its quadratic-penalty form separate over $\omega$ and $q$.
Applying the arguments of Sections~\ref{sec:equiv} and \ref{sec:pen} to each channel gives the optimal modes
\begin{equation}
\hat u_{k,q} = s_k \hat f_q \qquad(q = 1,\ldots,Q),\label{eq:membermv}
\end{equation}
where $s_k$ are given by \eqref{eq:member} for the constrained form and \eqref{eq:noise} for the quadratic-penalty form.
The memberships are the same for all channels because the distance $(\omega - \omega_k)^2$ does not depend on $q$.

Substituting the optimal modes \eqref{eq:membermv} into \eqref{eq:mvmd} and its quadratic-penalty form yields the unified reduced objective function
\begin{equation}
J_\gamma(\{\omega_k\}) = \sum_{\omega\in\Omega} \frac{P_{\mathrm{tot}}(\omega)}{\gamma + \sum_{k=1}^{K}(\omega - \omega_k)^{-2}},\label{eq:J_mvmd}
\end{equation}
where $P_{\mathrm{tot}}(\omega) = \sum_{q=1}^{Q}|\hat f_q(\omega)|^2$, with $\gamma = 0$ for the constrained form and $\gamma = 2\alpha$ for the quadratic-penalty form.
Setting the gradient of \eqref{eq:J_mvmd} to zero gives the center condition
\begin{equation}
\omega_k = \frac{\sum_{\omega} \omega P_{\mathrm{tot}}(\omega) s_k(\omega)^2}{\sum_{\omega} P_{\mathrm{tot}}(\omega) s_k(\omega)^2}.\label{eq:centermv}
\end{equation}
Thus, MVMD is equivalent to weighted FCM with the frequencies as the data points and the power summed over the channels as their weights.
The fixed-point conditions of ADMM derived in Section~\ref{sec:admm} also hold for MVMD with $P$ replaced by $P_{\mathrm{tot}}$.

\subsection{FCM-Based Algorithm for MVMD}\label{sec:mvmdfcm}
Since MVMD differs from VMD only in the weights of the corresponding FCM problem, the same FCM iteration applies to both.
Algorithm~\ref{alg:vmd} is therefore formulated for $Q$ channels, with the center update given by \eqref{eq:centermv}.

Computing $P_{\mathrm{tot}}$ before the iteration costs $O(QM)$, and constructing all mode spectra at the end costs $O(KQM)$.
Each FCM iteration costs $O(KM)$, independently of the number of channels $Q$.
In contrast, each ADMM iteration costs $O(KQM)$ because it updates the $KQ$ mode spectra.

\subsection{Numerical Experiments}\label{sec:mvmdexp}
We compare how the computing times of the FCM-based algorithm and ADMM for MVMD depend on the number of channels.

\emph{Test signals.} Each channel contains the AM--FM signal of Section~\ref{sec:vmdexp} with independently drawn phases and white Gaussian noise at an SNR of 20~dB.
We generate 20 realizations of a 1024-channel signal.
For each channel count $Q$, we use the first $Q$ channels of each realization.
The same inputs are used in Section~\ref{sec:smvmd}.

\emph{Solvers and measurement.} The signals are decomposed with $K = 3$ in the quadratic-penalty form using the solvers and stopping criterion of Section~\ref{sec:vmdexp}.
The computing time is measured over the same interval as in Section~\ref{sec:vmdexp}.
The computing time per iteration is also measured separately.

\begin{figure}[t]
\centering
\includegraphics{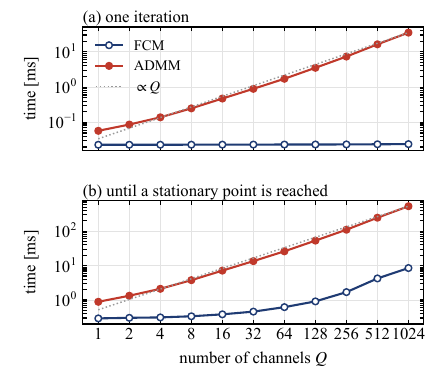}
\caption{Computing time of MVMD versus the number of channels $Q$ (AM--FM signal, SNR 20~dB, quadratic-penalty form; medians over 20 realizations).
(a) Computing time per iteration.
(b) Total computing time to reach a stationary point.
The dotted line is proportional to $Q$.}\label{fig:mvmd_timing}
\end{figure}

\emph{Computational cost.} Both solvers reached a stationary point for all 20 realizations at each $Q$.
The median iteration counts were 11 for the FCM-based algorithm and 15 for ADMM.
The normalized objective values reached by the two solvers differed by less than $10^{-15}$ in relative terms.

The computing time per FCM iteration remained nearly constant with $Q$, whereas that of ADMM increased with $Q$ (Fig.~\ref{fig:mvmd_timing}(a)), consistent with the computational costs derived in Section~\ref{sec:mvmdfcm}.
The total computing time of the FCM-based algorithm also increased for large $Q$ because the costs of computing $P_{\mathrm{tot}}$ and constructing the modes grow with $Q$.
Nevertheless, at $Q = 1024$, the total computing time was about 8.5~ms for the FCM-based algorithm, compared with 530~ms for ADMM (Fig.~\ref{fig:mvmd_timing}(b)).

\section{2D-VMD}
2D-VMD \cite{dragomiretskiy2015} extends VMD to images.
We show that 2D-VMD is also equivalent to weighted FCM and derive the corresponding FCM-based algorithm.

\subsection{Variational Problem and Equivalence to Weighted FCM}
Let $f(\bm{y})$ be a real-valued image with position $\bm{y} \in \mathbb R^2$, and let $\hat f(\bm{\omega}) = \int f(\bm{y}) e^{-j\bm{\omega}^\top \bm{y}} d\bm{y}$ be its two-dimensional Fourier transform.
The components of $\bm{\omega} \in \mathbb R^2$ are the horizontal and vertical angular frequencies.
Since $f$ is real-valued, $\hat f(-\bm{\omega})$ is the complex conjugate of $\hat f(\bm{\omega})$.
We call $\{\bm{\omega},-\bm{\omega}\}$ a conjugate pair.

2D-VMD \cite{dragomiretskiy2015} decomposes $f$ into $K$ real-valued modes $u_k$ with center frequencies $\bm{\omega}_k \in \mathbb R^2$ by solving
\begin{equation}
\min_{\{u_k\},\{\bm{\omega}_k\}} \sum_{k} \bigl\|\nabla\bigl[u_{\mathrm{AS},k}(\bm{y}) e^{-j\bm{\omega}_k^\top \bm{y}}\bigr]\bigr\|_2^2
\ \ \text{s.t.}\ \ \sum_k u_k = f,\label{eq:orig2d}
\end{equation}
where $u_{\mathrm{AS},k}$ is the two-dimensional analytic signal of $u_k$.
Whereas the one-dimensional analytic signal keeps the positive frequencies, $u_{\mathrm{AS},k}$ keeps the half-plane $\bm{\omega}^\top \bm{\omega}_k > 0$, with spectrum $\hat u_{\mathrm{AS},k} = (1 + \operatorname{sgn}(\bm{\omega}^\top \bm{\omega}_k)) \hat u_k$ \cite{dragomiretskiy2015}.
By Parseval's identity, each term of \eqref{eq:orig2d} is proportional to $\int\|\bm{\omega} - \bm{\omega}_k\|^2(1 + \operatorname{sgn}(\bm{\omega}^\top \bm{\omega}_k))^2|\hat u_k(\bm{\omega})|^2 d\bm{\omega}$.

Thus, of the two frequencies in each conjugate pair $\{\bm{\omega},\allowbreak-\bm{\omega}\}$, only the one nearer to $\bm{\omega}_k$ contributes to the bandwidth of mode $k$.
We define the squared distance from the conjugate pair to the center as
\begin{equation}
\begin{aligned}
d_k(\bm{\omega})^2 &= \min\bigl(\|\bm{\omega} - \bm{\omega}_k\|^2,\|\bm{\omega} + \bm{\omega}_k\|^2\bigr)\\
&= \|\bm{\omega}\|^2 + \|\bm{\omega}_k\|^2 - 2|\bm{\omega}^\top \bm{\omega}_k|,
\end{aligned}\label{eq:d2}
\end{equation}
which is the same for $\bm{\omega}$ and $-\bm{\omega}$.
Summing over one frequency of each conjugate pair and neglecting the line $\bm{\omega}^\top \bm{\omega}_k = 0$, which has measure zero, yields the frequency-domain form of \eqref{eq:orig2d}, up to a common positive factor:
\begin{equation}
\begin{aligned}
\min_{\{\hat u_k\},\{\bm{\omega}_k\}}\ & \sum_{\bm{\omega}\in\Omega_+} \sum_{k=1}^{K} d_k(\bm{\omega})^2|\hat u_k(\bm{\omega})|^2\\
\text{s.t.}\ \ & \sum_{k=1}^{K} \hat u_k(\bm{\omega}) = \hat f(\bm{\omega}) \quad(\forall\bm{\omega} \in \Omega_+).
\end{aligned}\label{eq:vmd2d}
\end{equation}
For a $T \times T$ image, the discrete Fourier transform gives the frequency grid $\bm{\omega} = (2\pi/T) \bm p$, where $\bm p \in \mathbb Z^2$ is the frequency index.
Let $\Omega_+$ contain one frequency of each nonzero conjugate pair, chosen as those with $\omega_2 > 0$, or $\omega_2 = 0$ and $\omega_1 > 0$, and let $M = |\Omega_+|$.
By conjugate symmetry, the reconstruction constraint need only be imposed on the frequencies in $\Omega_+$.

Problem \eqref{eq:vmd2d} is \eqref{eq:vmd} with $(\omega - \omega_k)^2$ replaced by $d_k(\bm{\omega})^2$.
The same correspondence holds for the quadratic-penalty form.
For each conjugate pair, the bandwidth and reconstruction error terms have factors of $4\alpha$ and $2$, respectively, yielding the same form as \eqref{eq:pen} with $\gamma = 2\alpha$.

For fixed centers, applying the arguments of Sections~\ref{sec:equiv} and \ref{sec:pen} with the distance $d_k(\bm{\omega})$ gives the optimal modes $\hat u_k = s_k \hat f$, where the memberships are given by
\begin{equation}
s_k = \frac{d_k^{-2}}{\gamma + \sum_{\ell} d_\ell^{-2}},\quad s_0 = \frac{\gamma}{\gamma + \sum_{\ell} d_\ell^{-2}} \quad(\gamma \ge 0).\label{eq:member2d}
\end{equation}
Substituting the optimal modes into \eqref{eq:vmd2d} and its quadratic-penalty form yields the unified reduced objective function
\begin{equation}
J_\gamma(\{\bm{\omega}_k\}) = \sum_{\bm{\omega}\in\Omega_+} \frac{P(\bm{\omega})}{\gamma + \sum_{k=1}^{K} d_k(\bm{\omega})^{-2}},\label{eq:J2d}
\end{equation}
where $P(\bm{\omega}) = |\hat f(\bm{\omega})|^2$, with $\gamma = 0$ for the constrained form and $\gamma = 2\alpha$ for the quadratic-penalty form.

Thus, 2D-VMD is equivalent to weighted FCM with $m = 2$, with the conjugate pairs as the data points, $P$ as their weights and $d_k$ as the distance to the center.
The quadratic-penalty form corresponds to FCM with a noise cluster at distance $\delta = 1/\sqrt\gamma$.

The modes at ADMM fixed points also satisfy the FCM membership condition in 2D-VMD.
In terms of conjugate pairs, the ADMM mode update \cite{dragomiretskiy2015} has the same form as the mode update in \eqref{eq:admm}, with the squared distance $(\omega - \omega_k)^2$ replaced by $d_k(\bm{\omega})^2$ defined in \eqref{eq:d2}.
Hence, the derivation of the membership condition in Section~\ref{sec:admm} applies to 2D-VMD as well.

Unlike in one-dimensional VMD, the optimal center for fixed memberships cannot be obtained directly as a weighted mean.
The objective function for each center is $F(\bm{\omega}_k) = \sum_{\bm{\omega}} P(\bm{\omega}) s_k(\bm{\omega})^2 d_k(\bm{\omega})^2$.
Because the nearer frequency of each conjugate pair depends on $\bm{\omega}_k$, $F$ is not quadratic in the center.
We therefore use the center update of 2D-VMD \cite{dragomiretskiy2015}.
With $\hat u_k = s_k \hat f$, this update can be expressed as
\begin{equation}
\bm{\omega}_k \leftarrow \frac{\sum_{\bm{\omega}} P(\bm{\omega}) s_k(\bm{\omega})^2 \operatorname{sgn}(\bm{\omega}^\top \bm{\omega}_k)\,\bm{\omega}}{\sum_{\bm{\omega}} P(\bm{\omega}) s_k(\bm{\omega})^2},\label{eq:fcmcenter2d}
\end{equation}
which computes the weighted mean of the frequencies of the conjugate pairs that are nearer to the current center.

Although the center update \eqref{eq:fcmcenter2d} does not necessarily minimize $F$, its fixed points satisfy the stationarity condition of the reduced objective function \eqref{eq:J2d} at differentiable points.
At a fixed point, the center remains unchanged by \eqref{eq:fcmcenter2d}.
Rearranging the resulting equality gives
\begin{equation}
\sum_{\bm{\omega}\in\Omega_+} P(\bm{\omega}) s_k(\bm{\omega})^2 \bigl(\operatorname{sgn}(\bm{\omega}^\top \bm{\omega}_k)\,\bm{\omega} - \bm{\omega}_k\bigr) = \bm 0.\label{eq:stat2d}
\end{equation}
Differentiating the squared distance \eqref{eq:d2} gives $\partial d_k^2/\partial\bm{\omega}_k = -2(\operatorname{sgn}(\bm{\omega}^\top \bm{\omega}_k)\,\bm{\omega} - \bm{\omega}_k)$ for $\bm{\omega}^\top \bm{\omega}_k \ne 0$.
Hence, at points where $J_\gamma$ is differentiable, \eqref{eq:stat2d} is equivalent to the stationarity condition $\nabla_{\bm{\omega}_k} J_\gamma = \bm 0$.
Since the original ADMM algorithm for 2D-VMD \cite{dragomiretskiy2015} uses the same center update, its fixed points also satisfy the stationarity condition at differentiable points of $J_\gamma$.

\subsection{FCM-Based Algorithm for 2D-VMD}\label{sec:2dfcm}
Based on the equivalence to weighted FCM, we construct an FCM-based algorithm for 2D-VMD, given as Algorithm~\ref{alg:2d}.
It alternately updates the memberships by \eqref{eq:member2d} and the centers by \eqref{eq:fcmcenter2d}.
It has the same structure as Algorithm~\ref{alg:vmd}, except for the squared distance defined in \eqref{eq:d2} and the center update in \eqref{eq:fcmcenter2d}.
Each iteration costs $O(KM)$.

Since $\bm{\omega}_k$ and $-\bm{\omega}_k$ represent the same center, the stationarity residual \eqref{eq:residual} is modified to $r = \max_k \min(\|\bar{\bm{\omega}}_k - \bm{\omega}_k\|,\|\bar{\bm{\omega}}_k + \bm{\omega}_k\|)/\Delta\omega$, where $\bar{\bm{\omega}}_k$ is the right-hand side of \eqref{eq:fcmcenter2d} and $\Delta\omega = 2\pi/T$.

The mode spectra are constructed only once, after the iteration terminates.

\begin{algorithm}[t]
  \DontPrintSemicolon
  \SetKwComment{Comment}{$\triangleright$\ }{}
  \SetCommentSty{textnormal}
  \caption{FCM-based algorithm for 2D-VMD}\label{alg:2d}
  \KwIn{spectrum $\hat f(\bm{\omega})$ of the image on $\Omega_+$, $K$, $\gamma \ge 0$, nonzero initial centers $\bm{\omega}_k \in \mathbb R^2$, tolerance $\varepsilon$, maximum iterations $N$}
  \KwOut{centers $\bm{\omega}_k$, modes $\hat u_k(\bm{\omega})$, residual $\hat u_0(\bm{\omega})$}
  \For{each $\bm{\omega} \in \Omega_+$}{
    $P(\bm{\omega}) \gets |\hat f(\bm{\omega})|^2$\;
  }
  \For{$i = 1,\ldots,N$}{
    \For(\Comment*[f]{memberships}){each $\bm{\omega} \in \Omega_+$}{
      \For{$k = 1,\ldots,K$}{
        $a_k \gets \|\bm{\omega}\|^2 + \|\bm{\omega}_k\|^2 - 2|\bm{\omega}^\top \bm{\omega}_k|$\;
      }
      \eIf{$\min_k a_k > 0$}{
        \For{$k = 1,\ldots,K$}{
          $s_k(\bm{\omega}) \gets a_k^{-1} \big/\bigl(\gamma + \sum_\ell a_\ell^{-1}\bigr)$\;
        }
        $s_0(\bm{\omega}) \gets \gamma\big/\bigl(\gamma + \sum_\ell a_\ell^{-1}\bigr)$\;
      }{
        $n_0 \gets$ number of $k$ with $a_k = 0$\;
        \For{$k = 1,\ldots,K$}{
          $s_k(\bm{\omega}) \gets [a_k = 0]/n_0$\;
        }
        $s_0(\bm{\omega}) \gets 0$\;
      }
    }
    \For(\Comment*[f]{centers}){$k = 1,\ldots,K$}{
      $\bar{\bm{\omega}}_k \gets \dfrac{\sum_{\bm{\omega}} \operatorname{sgn}(\bm{\omega}^\top \bm{\omega}_k)\,\bm{\omega} P(\bm{\omega}) s_k(\bm{\omega})^2}{\sum_{\bm{\omega}} P(\bm{\omega}) s_k(\bm{\omega})^2}$\;
    }
    $r \gets \max_k \min(\|\bar{\bm{\omega}}_k - \bm{\omega}_k\|,\|\bar{\bm{\omega}}_k + \bm{\omega}_k\|)/\Delta\omega$\;
    \lIf{$r < \varepsilon$ or $i = N$}{\textbf{break}}
    \For{$k = 1,\ldots,K$}{
      $\bm{\omega}_k \gets \bar{\bm{\omega}}_k$\;
    }
  }
  \For(\Comment*[f]{modes and residual}){$k = 0,\ldots,K$}{
    \For{each $\bm{\omega} \in \Omega_+$}{
      $\hat u_k(\bm{\omega}) \gets s_k(\bm{\omega}) \hat f(\bm{\omega})$\;
    }
  }
\end{algorithm}

\subsection{Numerical Experiments}
We compare the FCM-based algorithm and ADMM for 2D-VMD in terms of iteration counts, computing times and objective values.

\emph{Test image.} We use the same synthetic image \cite{gilles2012} as in the original 2D-VMD study \cite{dragomiretskiy2015}.
The image has $256 \times 256$ pixels ($T = 256$) and consists of a non-oscillating component comprising a filled ellipse and a rectangle, and four oscillating textures with different orientations and periods.
We remove the DC component and the Nyquist row and column from its spectrum.

\emph{Solvers and measurement.} We decompose the image into $K = 5$ modes in the constrained form ($\gamma = 0$) using Algorithm~\ref{alg:2d} and the original ADMM algorithm for 2D-VMD \cite{dragomiretskiy2015}.
We generate 100 sets of initial centers by uniform sampling from $\bm{\omega}_k/\Delta\omega \in [-T/2,T/2) \times [0,T/2)$.
Both solvers start from the same initial centers for each run.

As in Section~\ref{sec:vmdexp}, we tune both ADMM parameters $\alpha$ and $\tau$.
The candidate values of $\alpha$ are specified through \eqref{eq:delta} by $\delta$ ranging from about $11\Delta\omega$ to $114\Delta\omega$, and those of $\tau$ range from 2 to 3.5.
Among the settings for which all 100 runs reached a stationary point within 1000 iterations, we select $\delta \approx 36\Delta\omega$ and $\tau = 2.5$, which gave the smallest median iteration count.
The stopping criterion is the same as in Section~\ref{sec:vmdexp}, with the stationarity residual defined in Section~\ref{sec:2dfcm}.
The computing time is measured over the same interval as in Section~\ref{sec:vmdexp}.

\begin{figure}[!t]
\centering
\includegraphics[width=\textwidth]{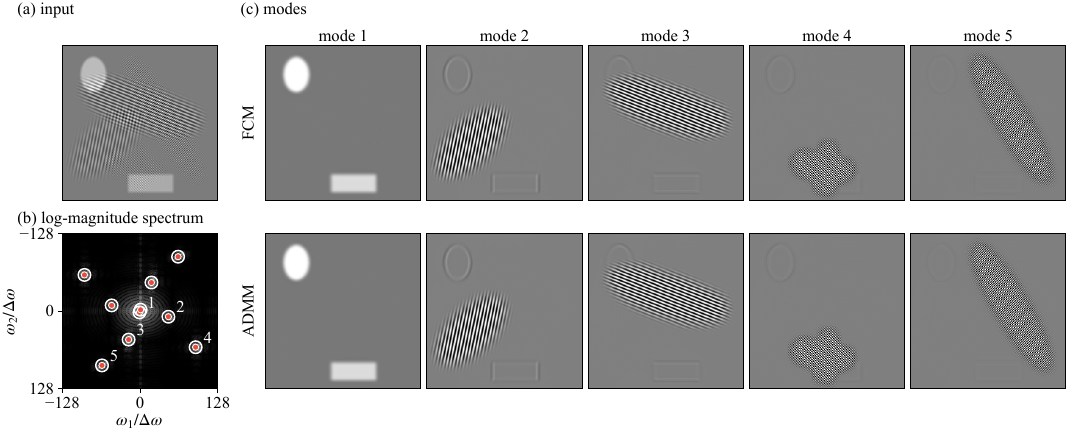}
\caption{Decomposition of the synthetic image \cite{gilles2012} by the FCM-based algorithm and ADMM from the same initial centers.
(a) Input image.
(b) Log-magnitude spectrum.
White circles: centers of the FCM-based algorithm.
Red dots: centers of ADMM (both at $\pm \bm{\omega}_k$, numbered by mode).
(c) Modes ordered by the norm of the center frequencies (common gray scale in each column).
Frequencies are in units of $\Delta\omega = 2\pi/T$, with the vertical frequency increasing downward.}\label{fig:texture}
\end{figure}

\begin{table}[t]
\caption{Median Iteration Counts, Computing Times and Normalized Objective Values for 2D-VMD (100 Initializations; Minimum and Maximum in Parentheses)}\label{tab:2d}
\centering
\begin{tabular}{@{}lrrr@{}}
\toprule
Solver & Iterations & Time [ms] & $J_0/\sum P$ [$(\Delta\omega)^2$]\\
\midrule
FCM & 11 (7--21) & 54 (35--104) & 6.307 (6.292--6.350)\\
ADMM & 27 (20--311) & 236 (174--2696) & 6.307 (6.292--6.350)\\
\bottomrule
\end{tabular}
\end{table}

\emph{Computational cost.} Table~\ref{tab:2d} compares the two solvers in terms of iteration counts, computing times and normalized objective values, reporting the medians with the minimum and maximum in parentheses.
Both solvers reached a stationary point from all 100 initializations.
The median iteration counts were 11 for the FCM-based algorithm and 27 for ADMM, and the median computing time of ADMM was about 4.3 times that of the FCM-based algorithm.

\emph{Objective values and modes.} The normalized objective values reached depended on the initial centers, but their median and range were the same for the two solvers.
Fig.~\ref{fig:texture} shows the decomposition results for one initialization.
The two solvers gave nearly identical modes, separating the non-oscillating component and the four oscillating textures.

\section{Successive MVMD}\label{sec:smvmd}
SMVMD \cite{liu2022a}, which combines successive VMD \cite{nazari2020} with MVMD, extracts the modes successively and does not require the number of modes to be given in advance.
We express the optimal mode for a fixed center using FCM memberships and derive an FCM-based algorithm.

\subsection{Variational Problem and Weighted FCM Memberships}
SMVMD extracts the $k$-th mode after the first $k - 1$ modes have been extracted.
It divides each channel $f_q$ into the $k$-th mode $u_{k,q}$ and the remaining signal $g_q = f_q - u_{k,q}$, which contains the $k - 1$ modes already extracted and the part of $f_q$ not yet extracted.
The mode and its center $\omega_k$ are determined by three requirements: (i) the mode has a small bandwidth around $\omega_k$; (ii) the remaining signal has little power near $\omega_k$, so that the mode captures the component around $\omega_k$; and (iii) the mode has little power near the centers $\omega_1,\ldots,\omega_{k-1}$ of the modes already extracted.
To enforce requirements (ii) and (iii), SMVMD uses filters of the form $B_{\alpha,\omega_k}$, whose frequency response $1/[\alpha(\omega - \omega_k)^2]$ is large near $\omega_k$, where $\alpha$ is the penalty factor.
SMVMD \cite{liu2022a} then solves
\begin{equation}
\begin{aligned}
\min_{\{u_{k,q},g_q\},\,\omega_k}\ \sum_{q=1}^Q
\Bigl( &\alpha\|\partial_t[u_{k,q}^{+}(t) e^{-j\omega_k t}]\|_2^2
+ \|B_{\alpha,\omega_k} g_q\|_2^2\\
&+ \sum_{\ell<k}\|B_{\alpha,\omega_\ell} u_{k,q}\|_2^2\Bigr)\\
\text{s.t.}\quad &u_{k,q} + g_q = f_q \quad(\forall q).
\end{aligned}\label{eq:sm_time}
\end{equation}

In the frequency domain, each term of \eqref{eq:sm_time} is a sum over the frequencies of $|\hat u_{k,q}|^2$ or $|\hat g_q|^2$ weighted by a function of $\omega$.
To express the SMVMD problem \cite{liu2022a} in a simpler form, we introduce the notation
\begin{equation}
\begin{gathered}
h_k(\omega) = \alpha^2(\omega - \omega_k)^4,\qquad
A_k(\omega) = 2\alpha(\omega - \omega_k)^2 + S_k(\omega),\\
S_k(\omega) = \sum_{\ell<k} \frac{1}{\alpha^2(\omega - \omega_\ell)^4}.
\end{gathered}\label{eq:sm_cost}
\end{equation}
Here $1/h_k$ is the squared magnitude of the frequency response of $B_{\alpha,\omega_k}$, and $S_k$ is the sum of the squared magnitudes of the frequency responses of the filters associated with the previously extracted modes.
The discrete counterpart of \eqref{eq:sm_time} is then
\begin{equation}
\begin{aligned}
\min_{\{\hat u_{k,q},\hat g_q\},\,\omega_k}\
&\sum_{q=1}^Q \sum_{\omega\in\Omega}
\Bigl(A_k(\omega)|\hat u_{k,q}(\omega)|^2 + \frac{|\hat g_q(\omega)|^2}{h_k(\omega)}\Bigr)\\
\text{s.t.}\ \ &\hat u_{k,q}(\omega) + \hat g_q(\omega) = \hat f_q(\omega) \quad(\forall\omega \in \Omega,\ \forall q),
\end{aligned}\label{eq:sm_allocation}
\end{equation}
which SMVMD solves by ADMM \cite{liu2022a}.

As in \eqref{eq:pen}, removing the constraint and penalizing the reconstruction error $\hat e_q = \hat f_q - \hat u_{k,q} - \hat g_q$ gives the quadratic-penalty form
\begin{equation}
\begin{aligned}
\min_{\{\hat u_{k,q},\hat g_q\},\,\omega_k}\
\sum_{q=1}^Q \sum_{\omega\in\Omega}
\Bigl( &A_k(\omega)|\hat u_{k,q}(\omega)|^2 + \frac{|\hat g_q(\omega)|^2}{h_k(\omega)}\\
&+ \frac{|\hat e_q(\omega)|^2}{\gamma}\Bigr),
\end{aligned}\label{eq:sm_pen}
\end{equation}
with $\gamma > 0$.
We use $\gamma = 0$ to denote the constrained form \eqref{eq:sm_allocation}.

For a fixed center and a fixed $\alpha$, \eqref{eq:sm_allocation} and \eqref{eq:sm_pen} separate over $\omega$ and $q$.
As in Section~\ref{sec:equiv}, the stationarity conditions are
\begin{equation}
A_k \hat u_{k,q} = \nu,\quad \hat g_q = h_k \nu,\quad \hat e_q = \gamma\nu,
\label{eq:sm_stat}
\end{equation}
where $\nu$ is the Lagrangian multiplier associated with the constraint $\hat u_{k,q} + \hat g_q + \hat e_q = \hat f_q$.
Since the three spectra sum to $\hat f_q$, \eqref{eq:sm_stat} gives the optimal spectra
\begin{equation}
\begin{gathered}
\hat u_{k,q} = s_k \hat f_q,\qquad \hat g_q = s_{\mathrm{c}} \hat f_q,\qquad \hat e_q = s_{\mathrm{e}} \hat f_q,\\
(s_k,s_{\mathrm{c}},s_{\mathrm{e}}) = \frac{(1,\,A_k h_k,\,\gamma A_k)}{1 + A_k(h_k + \gamma)} \qquad(\gamma \ge 0).
\end{gathered}\label{eq:sm_members}
\end{equation}
The memberships $s_k$, $s_{\mathrm{c}}$ and $s_{\mathrm{e}}$ are nonnegative, sum to one, and are the same for all channels.
Expressing the spectra in terms of the memberships as in \eqref{eq:sm_members}, the objective function of \eqref{eq:sm_pen} becomes
\begin{equation}
\sum_{\omega\in\Omega} P_{\mathrm{tot}}(\omega) \Bigl[A_k(\omega) s_k(\omega)^2 + \frac{s_{\mathrm{c}}(\omega)^2}{h_k(\omega)} + \frac{s_{\mathrm{e}}(\omega)^2}{\gamma}\Bigr],
\label{eq:sm_fcm}
\end{equation}
with $P_{\mathrm{tot}}$ as in \eqref{eq:J_mvmd}.
For the constrained form ($\gamma = 0$), $s_{\mathrm{e}} = 0$ and the last term is absent.

This has the same form as the objective of weighted FCM with $m = 2$.
As in MVMD, the data points are the frequencies $\omega$, and their weights are $P_{\mathrm{tot}}(\omega)$.
The mode and the remaining signal correspond to two clusters with squared distances $A_k$ and $1/h_k$, respectively.
For $\gamma > 0$, the reconstruction error corresponds to a noise cluster with squared distance $1/\gamma$.
For fixed $\omega_k$, the memberships in \eqref{eq:sm_members} coincide with the optimal FCM memberships.

Substituting the optimal memberships \eqref{eq:sm_members} into \eqref{eq:sm_fcm} gives the reduced objective function
\begin{equation}
J_{k,\gamma}(\omega_k) = \sum_{\omega\in\Omega} P_{\mathrm{tot}}(\omega)
\frac{A_k(\omega)}{1 + A_k(\omega)[h_k(\omega) + \gamma]}.
\label{eq:sm_reduced}
\end{equation}

For the center, we use the update of SMVMD \cite{liu2022a}, the spectral centroid of the mode.
With the optimal modes $\hat u_{k,q} = s_k \hat f_q$, the centroid becomes
\begin{equation}
\omega_k \leftarrow
\frac{\sum_{\omega\in\Omega} \omega P_{\mathrm{tot}}(\omega) s_k(\omega)^2}
{\sum_{\omega\in\Omega} P_{\mathrm{tot}}(\omega) s_k(\omega)^2},
\label{eq:sm_center}
\end{equation}
where $s_k$ on the right-hand side is evaluated at the current $\omega_k$.
Substituting the definition of $A_k$ in \eqref{eq:sm_cost} into \eqref{eq:sm_fcm}, we find that, for fixed memberships and $\alpha$, only two terms depend on $\omega_k$: $\sum_\omega P_{\mathrm{tot}}(\omega) s_k(\omega)^2 \cdot 2\alpha(\omega - \omega_k)^2$ and $\sum_\omega P_{\mathrm{tot}}(\omega) s_{\mathrm{c}}(\omega)^2/h_k(\omega)$.
The center update \eqref{eq:sm_center} has the form of \eqref{eq:center} and minimizes the first term but does not account for the second term.
Therefore, a fixed point of \eqref{eq:sm_center} is not in general a stationary point of the reduced objective function \eqref{eq:sm_reduced}.

\subsection{FCM-Based Algorithm for SMVMD}\label{sec:smfcm}
We construct an FCM-based algorithm for SMVMD, given as Algorithm~\ref{alg:smvmd}, by applying the optimal membership \eqref{eq:sm_members} to the successive mode extraction of SMVMD \cite{liu2022a}.

Algorithm~\ref{alg:smvmd} has an outer loop over the modes and an inner loop that determines each mode with $S_k$ fixed.
One iteration of the inner loop evaluates $h_k$ and $A_k$ of \eqref{eq:sm_cost} at the current center and $\alpha$, computes $s_k$ by \eqref{eq:sm_members}, and updates the center by \eqref{eq:sm_center}.
During the inner loop, $\alpha$ is gradually increased up to $\alpha_{\max}$, as in SMVMD \cite{liu2022a}.
The inner loop stops when the squared relative change of the mode spectra is at most $\varepsilon_s$ or after $N_s$ iterations.
Since $\sum_{\omega,q}|s_k(\omega) \hat f_q(\omega) - s_{\mathrm{old}}(\omega) \hat f_q(\omega)|^2 = \sum_\omega P_{\mathrm{tot}}(\omega)[s_k(\omega) - s_{\mathrm{old}}(\omega)]^2$, where $s_{\mathrm{old}}$ is the membership of the previous iteration, the squared relative change can be computed from the memberships without forming the modes.

Since the number of modes is not given in advance, the outer loop must decide when to stop extracting modes.
For this purpose, Algorithm~\ref{alg:smvmd} keeps the weight of the residual, $R = 1 - \sum_{\ell<k} s_\ell$, which is one at the start.
After the inner loop, $R$ is reduced by $s_k$. $S_{k+1}$ is then updated and remains fixed while the next mode is extracted.
The outer loop ends when the power of the residual spectra $R\hat f_q$, $\sum_\omega P_{\mathrm{tot}}(\omega) R(\omega)^2$, is at most a user-specified fraction $\rho$ of the total power, or when $K_{\max}$ modes have been extracted.
The mode spectra $s_k \hat f_q$ and the residual spectra $R\hat f_q$ are then constructed once, as in Algorithm~\ref{alg:vmd}.

\begin{algorithm}[!t]
  \DontPrintSemicolon
  \SetKwComment{Comment}{$\triangleright$\ }{}
  \SetKwComment{Note}{}{}
  \SetCommentSty{textnormal}
  \caption{FCM-based algorithm for SMVMD}\label{alg:smvmd}
  \KwIn{unilateral spectra $\hat f_q(\omega)$ ($q = 1,\ldots,Q$), $\Omega$, $\gamma \ge 0$, initial center $\omega_{\mathrm{init}}$, $\alpha_{\min}$, $\alpha_{\max}$, $\beta$, tolerances $\varepsilon_s$, $\rho$, maximum iterations $N_s$, maximum number of modes $K_{\max}$}
  \KwOut{number of modes $K$, centers $\omega_k$, modes $\hat u_{k,q}(\omega)$, residuals $\hat u_{0,q}(\omega)$}
  \For{each $\omega \in \Omega$}{
    $P_{\mathrm{tot}}(\omega) \gets \sum_{q=1}^Q|\hat f_q(\omega)|^2$\;
    $R(\omega) \gets 1$\;
    $S_1(\omega) \gets 0$\;
  }
  $K \gets 0$\;
  \While{$K < K_{\max}$ and $\sum_\omega P_{\mathrm{tot}}(\omega) R(\omega)^2 > \rho\sum_\omega P_{\mathrm{tot}}(\omega)$}{
    $k \gets K + 1$\;
    $\omega_k \gets \omega_{\mathrm{init}}$\;
    \For{each $\omega \in \Omega$}{
      $s_{\mathrm{old}}(\omega) \gets 0$\;
    }
    \For(\Comment*[f]{iterations for the $k$-th mode}){$i = 1,\ldots,N_s$}{
      $\alpha \gets \min(\alpha_{\min} \beta^{i-1},\alpha_{\max})$\;
      \For{each $\omega \in \Omega$}{
        $h_k(\omega) \gets \alpha^2(\omega - \omega_k)^4$\;
        $A_k(\omega) \gets 2\alpha(\omega - \omega_k)^2 + S_k(\omega)$\;
        $s_k(\omega) \gets 1\big/[1 + A_k(\omega)(h_k(\omega) + \gamma)]$\;
      }
      $\bar\omega_k \gets \dfrac{\sum_\omega \omega P_{\mathrm{tot}}(\omega) s_k(\omega)^2}
      {\sum_\omega P_{\mathrm{tot}}(\omega) s_k(\omega)^2}$\;
      $\kappa \gets \dfrac{\sum_\omega P_{\mathrm{tot}}(\omega)[s_k(\omega) - s_{\mathrm{old}}(\omega)]^2}{\sum_\omega P_{\mathrm{tot}}(\omega) s_{\mathrm{old}}(\omega)^2 + \epsilon_{\mathrm{mach}}}$\;
      \Note{($\epsilon_{\mathrm{mach}}$: machine epsilon)}
      \lIf{$\kappa \le \varepsilon_s$ or $i = N_s$}{\textbf{break}}
      $s_{\mathrm{old}} \gets s_k$;\ $\omega_k \gets \bar\omega_k$\;
    }
    $\alpha_k \gets \alpha$\;
    $K \gets k$\;
    \For{each $\omega \in \Omega$}{
      $R(\omega) \gets R(\omega) - s_k(\omega)$\;
      $S_{k+1}(\omega) \gets S_k(\omega) + 1\big/[\alpha_k^2(\omega - \omega_k)^4]$\;
    }
  }
  \For{each $\omega \in \Omega$}{
    $s_0(\omega) \gets R(\omega)$\;
  }
  \For(\Comment*[f]{modes and residuals}){$k = 0,\ldots,K$}{
    \For{each $\omega \in \Omega$}{
      \For{$q = 1,\ldots,Q$}{
        $\hat u_{k,q}(\omega) \gets s_k(\omega) \hat f_q(\omega)$\;
      }
    }
  }
\end{algorithm}

The ADMM iteration of SMVMD \cite{liu2022a} updates the mode, the center and the Lagrangian multiplier in turn.
The Lagrangian multiplier is updated by dual ascent with step size $\tau$.
For a fixed center and a fixed $\alpha$, the modes at the fixed points of the ADMM mode and multiplier updates coincide with the optimal modes in \eqref{eq:sm_members}.
The correspondence between $\tau$ and $\gamma$ is as follows: $\gamma = 1$ for $\tau = 0$ with the multiplier initialized to zero, and $\gamma = 0$ for $\tau > 0$.
Algorithm~\ref{alg:smvmd} computes these optimal modes directly through $s_k$.

Each iteration of Algorithm~\ref{alg:smvmd} costs $O(M)$, independently of $Q$, after the $O(QM)$ computation of $P_{\mathrm{tot}}$.
Constructing the $(K + 1) Q$ output spectra costs $O(KQM)$.
In contrast, each ADMM iteration of SMVMD costs $O(QM)$, because it updates the mode spectra of all $Q$ channels.

Sections~\ref{sec:vmd}--\ref{sec:smvmd} have related the four decompositions to weighted FCM in the same way: for fixed centers, the optimal modes are the input spectra multiplied by weighted FCM memberships.
Table~\ref{tab:corr} summarizes this correspondence and the resulting algorithms.

\begin{table}[!t]
\caption{Correspondence Between the Decompositions and Weighted FCM}\label{tab:corr}
\centering
\footnotesize
\setlength{\tabcolsep}{4pt}
\begin{tabular}{@{}>{\raggedright\arraybackslash}p{32mm}ll>{\raggedright\arraybackslash}p{25mm}>{\raggedright\arraybackslash}p{45mm}@{}}
\toprule
& VMD & MVMD & 2D-VMD & SMVMD ($k$-th mode)\\
\midrule
Data points & $\omega \in \Omega$ & $\omega \in \Omega$ & pairs $\{\bm{\omega},-\bm{\omega}\}$,\newline $\bm{\omega} \in \Omega_+$ & $\omega \in \Omega$\\
Weights & $P(\omega)$ & $P_{\mathrm{tot}}(\omega)$ & $P(\bm{\omega})$ & $P_{\mathrm{tot}}(\omega)$\\
Squared distances & $(\omega - \omega_k)^2$ & $(\omega - \omega_k)^2$ & $d_k(\bm{\omega})^2$ in \eqref{eq:d2} & $A_k$ (mode),\newline $1/h_k$ (remaining signal) in \eqref{eq:sm_cost}\\
Noise cluster ($\tau = 0$),\newline $\delta = 1/\sqrt\gamma$ & $\gamma = 2\alpha$ & $\gamma = 2\alpha$ & $\gamma = 2\alpha$ & $\gamma = 1$\\
Memberships & \eqref{eq:member}, \eqref{eq:noise} & \eqref{eq:member}, \eqref{eq:noise} & \eqref{eq:member2d} & \eqref{eq:sm_members}\\
Reduced objective & \eqref{eq:Jgamma} & \eqref{eq:J_mvmd} & \eqref{eq:J2d} & \eqref{eq:sm_reduced}\\
Center update & \eqref{eq:center} & \eqref{eq:centermv} & \eqref{eq:fcmcenter2d} & \eqref{eq:sm_center}\\
Algorithm & \ref{alg:vmd} ($Q = 1$) & \ref{alg:vmd} & \ref{alg:2d} & \ref{alg:smvmd}\\
\bottomrule
\end{tabular}
\end{table}

\subsection{Numerical Experiments}
We compare the FCM-based algorithm and ADMM for SMVMD in terms of the extracted modes and computational cost.

\emph{Test signals.} For the single-channel experiments, we use the sinusoids and the AM--FM signal of Section~\ref{sec:vmdexp}, with the same noise levels and 100 realizations for each case.
For the multichannel experiments, we use the test signals of Section~\ref{sec:mvmdexp} ($Q = 1,2,4,\ldots,1024$, 20 realizations).

\emph{Solvers and measurement.} Algorithm~\ref{alg:smvmd} and ADMM are applied in both the constrained and quadratic-penalty forms.
In the quadratic-penalty form, Algorithm~\ref{alg:smvmd} uses $\gamma = 1$ and ADMM uses $\tau = 0$.
In the constrained form, Algorithm~\ref{alg:smvmd} uses $\gamma = 0$.
We test several values of $\tau$ for ADMM in the range from 0.25 to 6 and select $\tau = 2.5$, which gave the smallest median number of iterations.

Both solvers use the same mirror extension as in Section~\ref{sec:vmdexp}.
Both solvers start the extraction of each mode from $\omega_{\mathrm{init}} = \pi/2$ and use $\alpha_{\min} = 10/(2\pi)^2$ and $\beta = 2$.
We use $\alpha_{\max} = 1/(2\delta^2) = 312.5/(2\pi)^2$ in both forms, with $\delta = 0.08\pi$~rad/sample (40~Hz) as in Section~\ref{sec:vmdexp}.
This relation follows from \eqref{eq:sm_members} for the first mode with $\gamma = 1$.
We set $\varepsilon_s = 10^{-7}$, $N_s = 99$, $K_{\max} = 5$, and $\rho = 0.02$ at 20~dB and 0.25 at 5~dB.
The computing time is measured over the same interval as in Section~\ref{sec:vmdexp}.

\begin{table}[tb]
\caption{Median Number of Modes, Component Errors, Total Iteration Counts and Computing Times for SMVMD}\label{tab:sm_accuracy}
\centering
\setlength{\tabcolsep}{2.2pt}
\begin{tabular}{@{}lllrrrrrr@{}}
\toprule
& & & & & \multicolumn{2}{c}{Iterations} & \multicolumn{2}{c}{Time [ms]}\\
\cmidrule(lr){6-7}\cmidrule(l){8-9}
Form & Signal & SNR & $K$ & Error [\%] & FCM & ADMM & FCM & ADMM\\
\midrule
Quadratic & Sinusoids & 20 dB & 3 & 8.05 & 21 & 29 & 0.62 & 0.82\\
penalty & Sinusoids & 5 dB & 3 & 31.83 & 25 & 31 & 0.70 & 0.87\\
& AM--FM & 20 dB & 3 & 9.73 & 22 & 29 & 0.64 & 0.82\\
& AM--FM & 5 dB & 3 & 32.28 & 27 & 32 & 0.74 & 0.89\\
\addlinespace
Constrained & Sinusoids & 20 dB & 3 & 8.43 & 22 & 37 & 0.64 & 1.79\\
& Sinusoids & 5 dB & 3 & 40.88 & 26.5 & 40.5 & 0.74 & 1.93\\
& AM--FM & 20 dB & 3 & 8.40 & 22 & 38 & 0.65 & 1.84\\
& AM--FM & 5 dB & 3 & 40.86 & 27 & 42 & 0.75 & 2.02\\
\bottomrule
\end{tabular}
\end{table}

\emph{Extracted modes.} Table~\ref{tab:sm_accuracy} gives the medians of the number of modes $K$, component error, total number of iterations over all modes and computing time.
The component error is the relative 2-norm error of the extracted modes with respect to the three noise-free components.
Both solvers extracted three modes in every realization, with the same median component errors.
To quantify the agreement between the modes obtained by the two solvers, we compute the relative difference as the 2-norm of their difference divided by the 2-norm of the ADMM modes.
The median relative difference ranged from $2.8 \times 10^{-5}$ to $3.9 \times 10^{-5}$ in the quadratic-penalty form and from $6 \times 10^{-5}$ to $1 \times 10^{-4}$ in the constrained form.

\begin{figure}[t]
\centering
\includegraphics{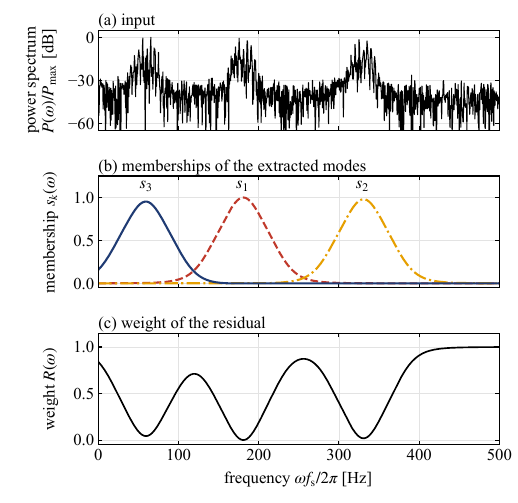}
\caption{Memberships obtained by the FCM-based algorithm for the AM--FM signal (SNR 20~dB, quadratic-penalty form).
(a) Input power spectrum $P(\omega)$ relative to its maximum $P_{\max}$.
(b) Memberships $s_k(\omega)$ of the extracted modes, numbered in the order of extraction.
(c) Residual weight $R(\omega)$.}\label{fig:sm_memberships}
\end{figure}

Fig.~\ref{fig:sm_memberships} shows the memberships for the AM--FM realization of Fig.~\ref{fig:membership}.
The memberships clearly show how the three signal components are assigned to the extracted modes, while the residual weight $R$ shows which frequencies remain in the residual.

\emph{Computational cost.} The median iteration count was smaller for the FCM-based algorithm than for ADMM in each of the eight cases (Table~\ref{tab:sm_accuracy}).
For these single-channel signals, ADMM took 1.2--1.3 times as long in the quadratic-penalty form and 2.6--2.8 times in the constrained form.

\begin{figure}[t]
\centering
\includegraphics{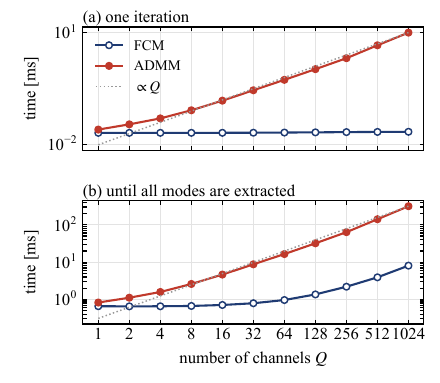}
\caption{Computing time of SMVMD versus the number of channels $Q$ (AM--FM signal, SNR 20~dB, quadratic-penalty form; medians over 20 realizations).
(a) Computing time per iteration.
(b) Total computing time to extract all modes.
The dotted line is proportional to $Q$.}
\label{fig:sm_timing}
\end{figure}

\emph{Multichannel signals.} To examine how the computational cost scales with the number of channels, we apply both solvers in the quadratic-penalty form with the same parameters as in the single-channel experiments.
The computing time per iteration and the total computing time are measured as in Section~\ref{sec:mvmdexp}.

Both solvers extracted three modes for all 20 realizations at each $Q$.
The FCM-based algorithm needed fewer iterations in total than ADMM for all realizations at each $Q$.
The median iteration counts were 21 for the FCM-based algorithm and 29 for ADMM.
The largest relative difference between the modes of the two solvers, computed over all channels with the modes in the order of extraction, was $4.7 \times 10^{-5}$.
The computing time per iteration of the FCM-based algorithm remained nearly constant with $Q$, whereas that of ADMM increased with $Q$ (Fig.~\ref{fig:sm_timing}(a)), consistent with the computational costs $O(M)$ and $O(QM)$ in Section~\ref{sec:smfcm}.
At $Q = 1024$, the total computing time was about 8.0~ms for the FCM-based algorithm and about 320~ms for ADMM (Fig.~\ref{fig:sm_timing}(b)).

\FloatBarrier
\section{Conclusion}
We showed that the constrained variational problem of VMD is equivalent to weighted FCM with $m = 2$, where the frequencies are the data points and the power spectrum gives their weights.
Its quadratic-penalty form corresponds to FCM with a noise cluster at distance $\delta = 1/\sqrt{2\alpha}$.
The ADMM fixed-point conditions coincide with the corresponding FCM conditions.
In the constrained form, the ADMM fixed-point conditions for the modes and centers are independent of $\alpha$ and $\tau$.
The equivalence allows VMD to be computed by an FCM-based algorithm that requires no dual-ascent parameter $\tau$ and requires $\alpha$ only in the quadratic-penalty form.
The noise distance gives a direct interpretation of $\alpha$ and can guide its selection based on the bandwidth of the components.
In the numerical experiments, the FCM-based algorithm needed fewer iterations and less computing time than ADMM even with tuned parameters and reached stationary points of comparable objective value.

We extended the equivalence to MVMD and 2D-VMD and derived their FCM-based algorithms.
In MVMD, the power summed over the channels gives the weights of the data points.
In 2D-VMD, pairs of conjugate frequencies serve as the data points.
For SMVMD, which extracts modes successively, we showed that the optimal mode for a fixed center at each extraction step can be expressed using FCM memberships.
Based on this correspondence, we derived an FCM-based algorithm for SMVMD.
In the numerical experiments, the FCM-based algorithms achieved results comparable to those of ADMM with shorter computing times.
For both MVMD and SMVMD, the computational cost per iteration of the FCM-based algorithm is independent of the number of channels.

By establishing a connection between VMD and FCM, this work opens the possibility of applying the theory and methods developed for FCM to VMD.

\section*{Acknowledgments}
This work was supported by JST PRESTO under Grant JPMJPR23O2, Japan.

\bibliographystyle{unsrt}
\bibliography{refs}

\end{document}